%% file: article.tex
\documentclass[conference]{IEEEtran}
\IEEEoverridecommandlockouts

\usepackage{url}        
\usepackage{booktabs}
\usepackage[colorlinks=true, linkcolor=blue, citecolor=blue, urlcolor=blue]{hyperref}

\usepackage{soul}
\usepackage{cite}
\usepackage{amsmath,amssymb,amsfonts}
\usepackage{algorithmic}
\usepackage{graphicx}
\usepackage{textcomp}
\usepackage{xcolor}
\def\BibTeX{{\rm B\kern-.05em{\sc i\kern-.025em b}\kern-.08em
    T\kern-.1667em\lower.7ex\hbox{E}\kern-.125emX}}

\begin{document}
\title{Deep Learning-Based Digitization of Overlapping ECG Images with Open-Source Python Code}

\author{
\IEEEauthorblockN{Reza Karbasi\textsuperscript{1$\dagger$}, Masoud Rahimi\textsuperscript{2$\dagger$}, Abdol-Hossein Vahabie\textsuperscript{3}, Hadi Moradi\textsuperscript{4}}\\
\IEEEauthorblockA{School of Electrical and Computer Engineering, University of Tehran, Tehran, Iran}\\
\IEEEauthorblockA{
    \textsuperscript{1}rezakarbasi@ut.ac.ir, 
    \textsuperscript{2}mr.rahimi39@ut.ac.ir, 
    \textsuperscript{3}h.vahabie@ut.ac.ir,
    \textsuperscript{4}moradih@ut.ac.ir}
\thanks{\textsuperscript{$\dagger$}These authors contributed equally to this work.}
}

\date{Jun 2025}

\maketitle

\begin{abstract}
\input{abstract.tex}
\end{abstract}


\section{Introduction}
\input{intro.tex}

\section{Related Works}
\input{related.tex}

\section{Materials and Methods}
\label{sec:method}
\input{method.tex}

\section{Experiments and Results}
\label{sec:results}
\input{results.tex}

\section{Discussion}
\input{discussion.tex}

\section{Conclusion}

\input{conclusion.tex}


\bibliographystyle{ieeetr}
\bibliography{ref}
\end{document}

%% file: abstract.tex
This paper addresses the persistent challenge of accurately digitizing paper-based electrocardiogram (ECG) recordings, with a particular focus on robustly handling single leads compromised by signal overlaps---a common yet under-addressed issue in existing methodologies. We propose a two-stage pipeline designed to overcome this limitation. The first stage employs a U-Net based segmentation network, trained on a dataset enriched with overlapping signals and fortified with custom data augmentations, to accurately isolate the primary ECG trace. The subsequent stage converts this refined binary mask into a time-series signal using established digitization techniques, enhanced by an adaptive grid detection module for improved versatility across different ECG formats and scales. Our experimental results demonstrate the efficacy of our approach. The U-Net architecture achieves an Intersection over Union (IoU) of 0.87 for the fine-grained segmentation task. Crucially, our proposed digitization method yields superior performance compared to a well-established baseline technique across both non-overlapping and challenging overlapping ECG samples. For non-overlapping signals, our method achieved a Mean Squared Error (MSE) of 0.0010 and a Pearson Correlation Coefficient ($\rho$) of 0.9644, compared to 0.0015 and 0.9366, respectively, for the baseline. On samples with signal overlap, our method achieved an MSE of 0.0029 and a $\rho$ of 0.9641, significantly improving upon the baseline's 0.0178 and 0.8676. This work demonstrates an effective strategy to significantly enhance digitization accuracy, especially in the presence of signal overlaps, thereby laying a strong foundation for the reliable conversion of analog ECG records into analyzable digital data for contemporary research and clinical applications. The implementation is publicly available at this GitHub repository: \href{https://github.com/masoudrahimi39/ECG-code}{https://github.com/masoudrahimi39/ECG-code}.

%% file: intro.tex
Electrocardiogram (ECG) serves as a cornerstone in the diagnosis and ongoing monitoring of cardiovascular diseases, which persist as a primary cause of mortality globally \cite{birnbaum2014role}. The ability to access and analyze ECG time-series data substantially enhances the efficacy of deep learning–based clinical decision support systems \cite{adedinsewo2022digitizing}. For many years, healthcare institutions have archived ECG records predominantly in paper or scanned image formats. These legacy records encapsulate invaluable clinical information, including detailed patient histories and documentation of rare cardiac events, crucial for longitudinal studies and personalized medicine \cite{reyna2024ecg}. In numerous hospitals, the practice of storing ECGs as images or PDF documents is prevalent, largely due to reduced costs and the obviation of requirements for specialized signal acquisition hardware or highly trained personnel for managing digital signal databases \cite{lence2023automatic}. Consequently, the task of digitizing these ECG images into structured, analyzable time-series data has emerged as an essential endeavor, promising to unlock a wealth of historical medical data for improved patient care, early detection of cardiac abnormalities, and advanced biomedical research.

Research in this domain pursues many aims; here, we focus on \emph{digitization}---recovering sample level signal values from legacy ECG images and rendering them as precise time series data for current clinical and research applications.
Methodologically, approaches are broadly categorized into classical computer vision techniques \cite{article-12} and more contemporary machine learning or deep learning-based strategies \cite{article-1, article-22}. Classical methods typically involve a sequence of image processing steps such as deskewing, noise reduction, grid processing, and waveform extraction using techniques like edge detection, morphological operations, or adaptive line tracking \cite{article-12, article-20, article-14, article-11}. While these can be robust to minor variations \cite{article-23}, their performance often hinges on meticulous parameter tuning and can degrade with significant variations in image quality or noise \cite{article-19}. In recent years, deep learning approaches, often employing architectures like U-Nets \cite{article-1, article-22, article-5, article-7} or Generative Adversarial Networks (GANs) \cite{article-15}, have shown considerable promise by learning complex features directly from image data. However, a persistent and critical challenge, particularly for compact multi-lead ECG formats, is the accurate reconstruction of signals when traces from different leads graphically overlap \cite{article-21, article-20}. Many existing methods, spanning both classical and deep learning techniques, either implicitly assume non-overlapping signals or acknowledge that overlapping traces lead to performance degradation \cite{article-13}. While some recent deep learning methods have begun to address this \cite{article-22}, the robust handling of signal overlaps remains a key area for improvement.

In this work, we address digitization of ovelapping signals by introducing a robust pipeline for digitizing single-lead paper ECGs, with a particular emphasis on effectively addressing the challenge of overlapping signals. Our proposed methodology comprises a two-stage process. The initial stage employs a deep learning-based segmentation network, specifically a U-Net architecture. The network trained on a dataset deliberately curated to include samples with signal overlap and augmented with custom techniques like OverlaySignal to enhance its resilience to such scenarios. The dataset\cite{dataset-paper} used for training our network was specifically prepared to include a high prevalence of samples with signal overlap. In the end, this stage accurately isolates the primary ECG trace from its background and any interfering elements. In the subsequent stage, the refined binary mask generated by the segmentation network is converted into a time-series signal using established signal processing techniques. This digitization is importantly preceded by an automated grid detection module, ensuring the adaptability of our process to various ECG paper scales and scanning resolutions. This work presents a method that significantly enhances the accuracy of ECG digitization, specifically for images with overlapping signals. Our approach addresses a key limitation of prior influential work, including the method by Wu et al. \cite{article-21}, which reports reduced performance when signals graphically interfere.





%% file: related.tex
The digitization of ECG signals from heterogeneous image sources is an established yet dynamic research field, fundamental for turning legacy paper archives into data amenable to algorithmic scrutiny across capture modalities. Objectives differ: some projects diagnose straight from ECG images (Yu et al.\,\cite{article-5}), others perform classification after reconstructing the waveform (Mishra et al.\,\cite{article-13}; Patil \& Karandikar\,\cite{article-16}; Isabel et al.\,\cite{article-8}), whereas our own goal—and that of many others—is the high-fidelity recovery of the one-dimensional signal itself. Input layouts span 12 × 1 (Lobodzinski et al.\,\cite{article-11}; Badilini et al.\,\cite{article-3}), 6 × 2 (Wu et al.\,\cite{article-22}) and 3 × 4 (Badilini et al.\,\cite{article-3}), arriving as scanned sheets (Li et al.\,\cite{article-1}; Fortune et al.\,\cite{article-23}; Wu et al.\,\cite{article-21}), smartphone photographs (Mishra et al.\,\cite{article-13}; Patil et al.\,\cite{article-15}; Isabel et al.\,\cite{article-8}) or synthetically generated images for network training (Yu et al.\,\cite{article-5}; Verlyck et al.\,\cite{article-6}; Silva et al.\,\cite{article-7}). Demolder et al.\cite{article-22} and Mallawaarachchi et al. \cite{article-12} highlight—and our own experiments confirm—that reconstruction accuracy is acutely sensitive to image quality, especially when photographs are taken with phone cameras. On the other hand, from the perspective of methodology, the literature divides into classical computer-vision pipelines (Mallawaarachchi et al.\,\cite{article-12}) and modern deep-learning systems (Li et al.\,\cite{article-1}; Demolder et al.\,\cite{article-22}) which will be discussed in the following.

Classical computer vision techniques for ECG digitization typically involve a sequential pipeline of image processing steps: pre-processing such as deskewing (Mallawaarachchi et al. \cite{article-12}, Wang et al. \cite{article-20}, Garg et al. \cite{article-9}) and noise reduction, followed by grid detection and removal, and culminating in waveform extraction using methods like edge detection (Mallawaarachchi et al. \cite{article-12}, Wang et al. \cite{article-20}), morphological operations (Mallawaarachchi et al. \cite{article-12}, Patil and Karandikar \cite{article-14}, Wang et al. \cite{article-20}), adaptive line tracking (Lobodzinski et al. \cite{article-11}), or entropy-based bit-plane slicing (Patil and Karandikar \cite{article-14}). These methods, exemplified by the work of Fortune et al. \cite{article-23} which achieved high correlation (0.977) using a classical pipeline, can be robust to minor image variations and do not require extensive training data. However, their performance often depends on careful parameter tuning and can be susceptible to significant variations in image quality, grid patterns, and noise levels (Waits and Soliman \cite{article-19}). For instance, while traditional methods might handle trace-grid overlap effectively (Lobodzinski et al. \cite{article-11}), they generally lack mechanisms to resolve overlapping ECG traces from different leads and are often evaluated on single, clearly separated waveforms (Mallawaarachchi et al. \cite{article-12}, Tabassum and Ahmad \cite{article-18}).

In recent years, machine learning and particularly deep learning approaches have demonstrated significant promise in ECG digitization. These methods, often leveraging architectures like U-Nets (Li et al. \cite{article-1}, Demolder et al. \cite{article-22}, Yu et al. \cite{article-5}, Silva et al. \cite{article-7}), Generative Adversarial Networks (GANs) (Patil et al. \cite{article-15}), and object detection models like YOLO for lead and its text localization (Patil et al. \cite{article-15}, Yu et al. \cite{article-5}, Verlyck et al. \cite{article-6}), can learn complex features directly from image data such as the placement of a lead signal or text. Studies report high performance on various datasets, including those with considerable noise (Li et al. \cite{article-1}) or those generated synthetically to cover a wide range of artifacts (Verlyck et al. \cite{article-6}, Yu et al. \cite{article-5}). Evaluation metrics such as Dice coefficient for segmentation (Li et al. \cite{article-1}), Pearson Correlation Coefficient (PCC), and Root Mean Squared Error (RMSE) (Demolder et al. \cite{article-22}) indicate strong concordance with ground truth signals under specific conditions. However, these models typically require large annotated datasets for optimal training, demand substantial computational resources, and their generalization to out-of-distribution data or highly degraded inputs can be challenging. 

A persistent and critical challenge in the digitization of electrocardiograms (ECGs), particularly pronounced for compact multi-lead formats such as the 3x4 layout, is the accurate reconstruction of signals when traces from different leads graphically overlap. A substantial body of existing literature, spanning both classical computer vision and numerous deep learning methodologies, often does not explicitly confront this intricate issue, frequently proceeding under the implicit assumption of non-overlapping signals. In instances where the problem of overlapping traces is acknowledged, it is commonly identified as an unresolved technical hurdle or a direct cause of significant performance degradation \cite{article-21, article-20}. For example, the otherwise highly automated and notably accurate digitization tool developed by Wu et al.\ \cite{article-21} exhibits a marked reduction in signal correlation, dropping to the 60--70\% range when processing overlapping leads, which contrasts sharply with the 97--99\% correlation achieved on non-overlapping data segments. Similarly, while the work by Fortune et al.\ \cite{article-23} details a robust classical digitization pipeline, its principal design does not focus on the resolution of such graphical interferences; furthermore, inherent limitations in handling sharp waveform features or images of compromised quality suggest that complex scenarios, such as dense signal overlaps, would present considerable difficulties. Some published methods offer only rudimentary strategies for handling overlaps, including the averaging of pixel positions in conflicted regions \cite{article-20}. 
Although the recent work of Demolder \emph{et al.}\,\cite{article-22} explicitly tackles signal‐overlap and reports a high peak correlation coefficient (PCC~$>0.91$), their framework still fails on roughly 6.6\% of ECGs that combine overlap with severe distortion and low-resolution scanning. This non-negligible failure rate confirms that reliably and consistently disentangling superimposed ECG traces remains an open challenge. Consequently, many state-of-the-art digitization pipelines—including several sophisticated approaches—continue to exhibit limitations when confronted with such complex inputs\,\cite{article-21}, underscoring the clear and ongoing need for dedicated, robust mechanisms for overlap resolution across diverse real-world conditions.

%% file: method.tex
The overall methodology for digitizing ECG images involves a two-stage process. Initially, a segmentation network based on the U-Net architecture is employed to extract the ECG signal mask from the input image of a single lead. Subsequently, a grid detection and signal digitization procedure is applied to convert this mask into a time-series data which will be explained int the following sections. 

\subsection{Dataset}


The models in this study were trained and evaluated using a suite of datasets generated from a new open-source framework designed to produce synthetic ECG images for various deep learning tasks. This framework provides four distinct datasets: (1) a \textbf{digitization} dataset, which pairs ECG images with their corresponding time-series signals; (2) a \textbf{detection} dataset, featuring YOLO-format annotations for localizing lead regions and names; (3) a \textbf{segmentation} dataset, comprising cropped single-lead images with pixel-level masks; and (4) an \textbf{overlap} dataset, designed specifically for signal interference challenges. A key feature of the overlap dataset is that its images contain superimposed waveforms from adjacent leads, while the ground truth masks remain "clean," isolating only the target waveform.

\begin{figure}[t]
  \centering
  \includegraphics[width=0.9\linewidth]{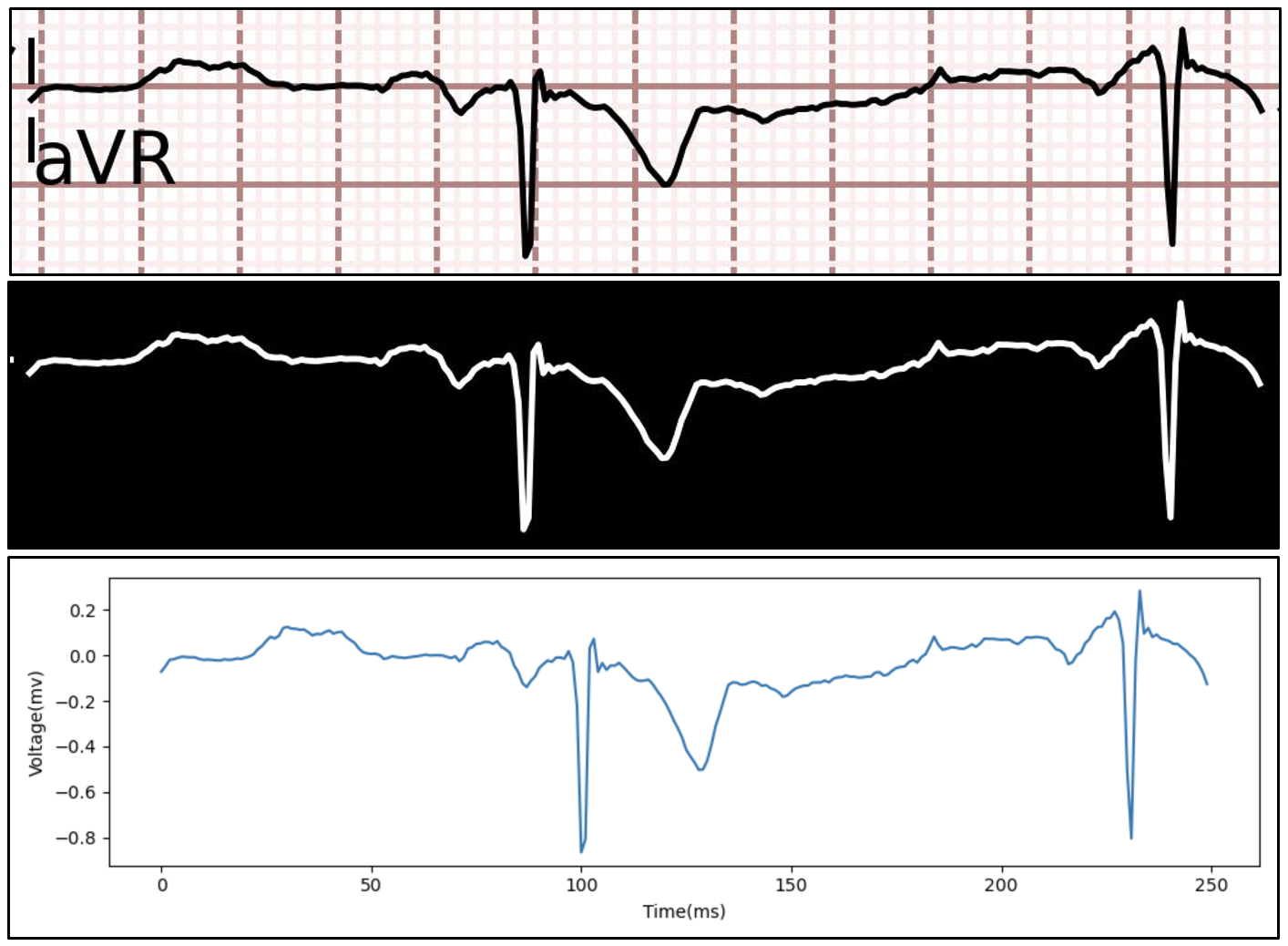} 
  \caption{An example of a raw ECG lead image from the dataset used. Upper, the input image of the pipeline. Middle image, mask in PNG/BMP format. Bottom, plot of the original image.}
  \label{fig:raw_dataset_sample}
\end{figure}

For the scope of this paper, we utilized the `segmentation` and `overlap` datasets from this suite. The segmentation model was trained on the `segmentation` dataset, which provides individual lead images (PNG), their corresponding ground truth masks (in both PNG and BMP formats for efficiency), and the original time-series signal (JSON). For evaluating the final pipeline and comparing it against the baseline method, we used the `overlap` dataset, which is specifically curated to contain 100 samples with manually verified signal overlaps and 185 non-overlapping samples. Figure~\ref{fig:raw_dataset_sample} displays a representative sample from the `segmentation` dataset used for training our model.

\subsection{ECG Signal Segmentation}
\label{sec:method-signal-segmentation}
The segmentation stage aims to accurately delineate the ECG waveform from the other parts of the image.

\subsubsection{Model Architecture and Training}
We employed a U-Net architecture \cite{Ronneberger2015} with a ResNet34 \cite{He2016} backbone, pre-trained on ImageNet, as our segmentation model. The model was configured to accept 3-channel RGB images as input and produce a single-channel binary mask as output. To expedite the development and training process, we utilized the \texttt{segmentation-models-pytorch} library \cite{Iakubovskii-2019}. The training pipeline was managed using PyTorch Lightning \cite{Falcon_PyTorch_Lightning_2019}.

The model was trained for 20 epochs with a batch size of 16. We used the Adam optimizer \cite{adam-optimizer} with a learning rate of $2 \times 10^{-4}$. The loss function was a weighted combination of Binary Cross-Entropy (BCE) loss and Dice loss, formulated as:
$$ \mathcal{L}_{\text{seg}} = \mathcal{L}_{\text{Dice}} + 0.5 \cdot \mathcal{L}_{\text{BCE}} $$
where
$$ \mathcal{L}_{\text{BCE}}(y, \hat{y}) = -[y \log(\hat{y}) + (1-y) \log(1-\hat{y})] $$
and the Dice loss for a binary case is
$$ \mathcal{L}_{\text{Dice}}(Y, \hat{Y}) = 1 - \frac{2 |Y \cap \hat{Y}|}{|Y| + |\hat{Y}|} $$
where $y$ is the ground truth pixel value, $\hat{y}$ is the predicted pixel value, $Y$ is the set of ground truth pixels, and $\hat{Y}$ is the set of predicted pixels.

\subsubsection{Data Preprocessing and Augmentation}
\begin{figure}[t]
  \centering
  \includegraphics[width=0.9\linewidth]{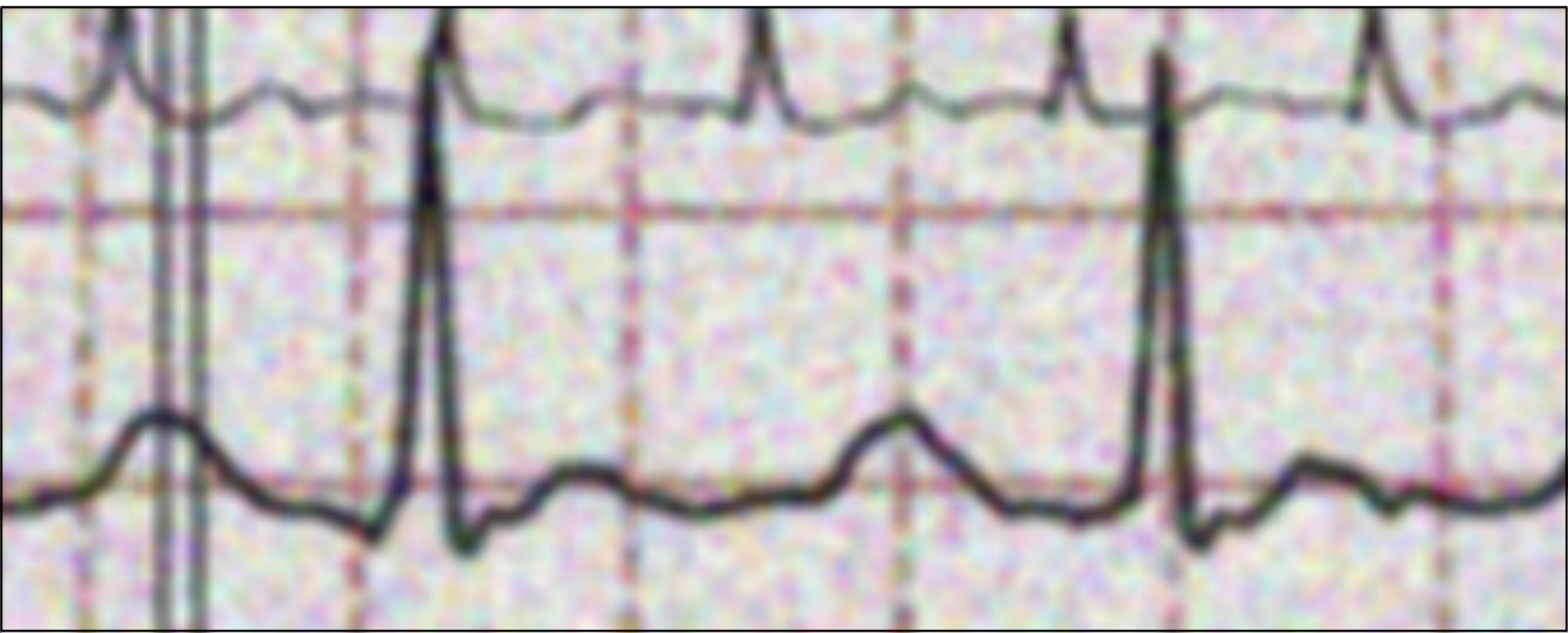} 
  \\[6pt]
  \includegraphics[width=0.9\linewidth]{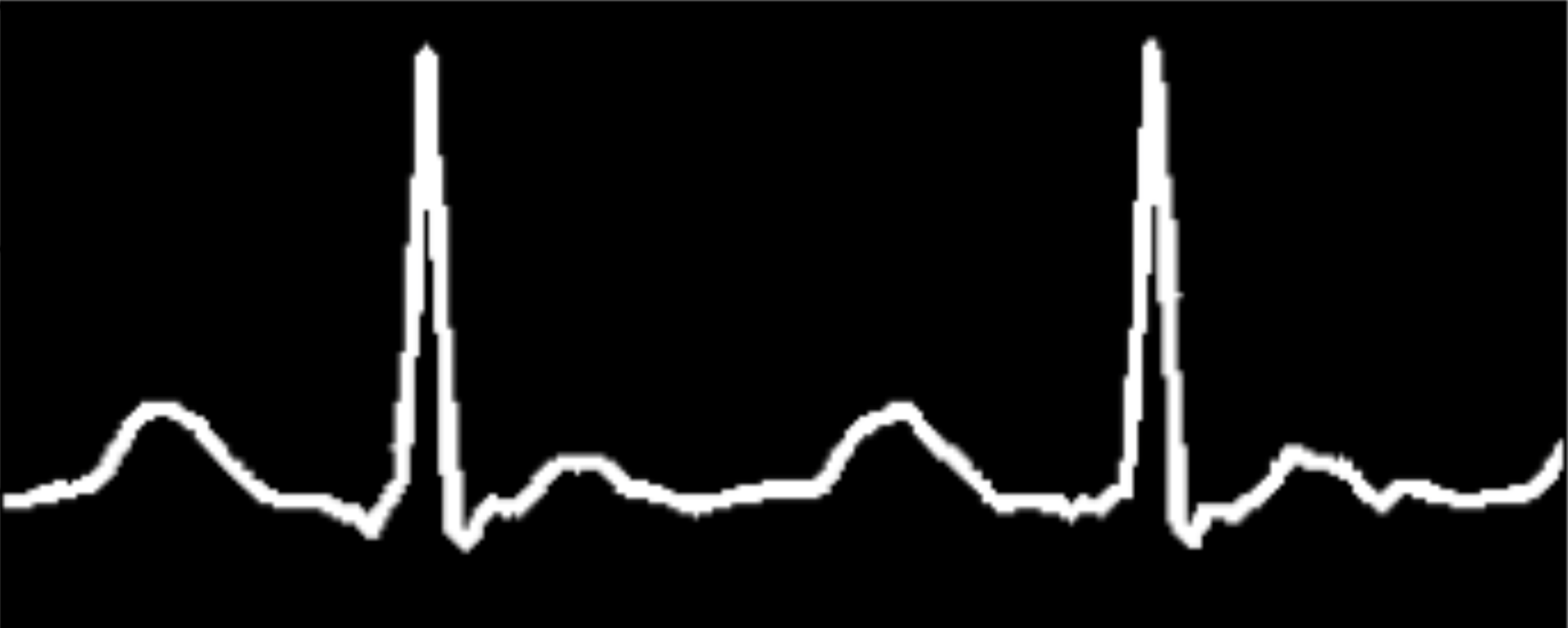} 
  \caption{Example of an input ECG lead image (upper) and the corresponding predicted segmentation mask (bottom) from the U-Net model.}
  \label{fig:segmentation_io}
\end{figure}

Due to the considerable length variation in some ECG leads, all input images and their corresponding masks were resized to a fixed dimension of $96 \times 960$ pixels (height $\times$ width) prior to training.
A series of data augmentation techniques were applied to enhance the model's robustness and generalization capabilities. The augmentation pipeline included:
\begin{itemize}
    \item Vertical Flip (probability $p=0.5$)
    \item Horizontal Flip ($p=0.5$)
    \item Custom OverlaySignal: Randomly overlays a segment of another ECG signal onto the top or bottom of the input image ($p=0.4$, crop height 30 pixels). This helps the model become robust to signal overlaps. (See the upper image in Figure~\ref{fig:overlay_signal_aug})
    \item Rotate (limit $\pm 5$ degrees, $p=0.3$)
    \item AddVerticalLines: Adds a few random vertical black lines across the image ($p=0.3$, 2 lines, 3 pixels width). This simulates certain types of noise or artifacts. (See the bottom image in Figure~\ref{fig:add_vertical_lines_aug})
    \item Gaussian Noise (variance limit $1.0-4.0$, $p=0.2$)
    \item Gaussian Blur (blur limit $1-5$ pixels, $p=0.2$)
\end{itemize}
The custom augmentations, \texttt{OverlaySignal} and \texttt{AddVerticalLines}, were specifically designed to improve the model's performance in the presence of signal overlaps and line-like artifacts, respectively. Figure~\ref{fig:segmentation_io} illustrates an example input image and the corresponding output mask generated for the training of the segmentation network. A key aspect of this training strategy is that the ground truth mask is always kept 'clean'; it exclusively highlights the single, target ECG waveform, cleanly separating it from background elements like grid lines and from any interfering signals.

\begin{figure}[t]
  \centering
  \includegraphics[width=0.9\linewidth]{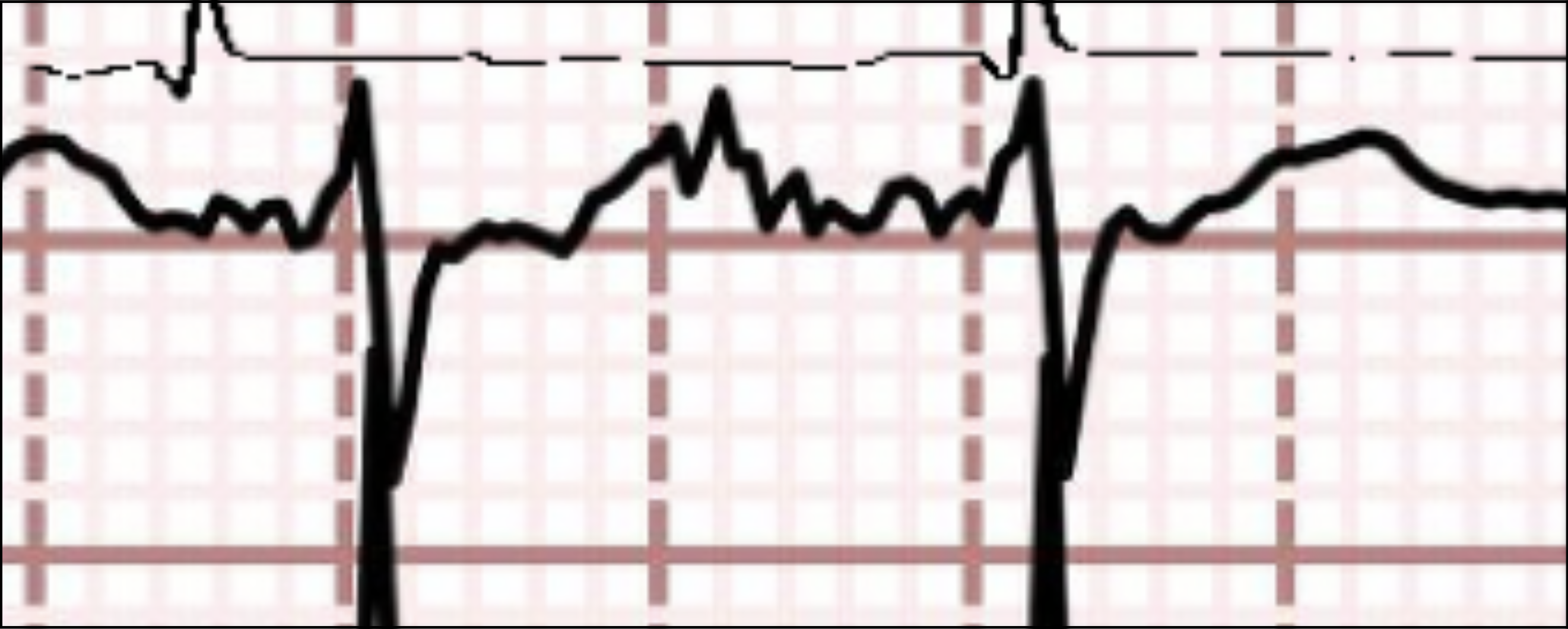} 
  \\[6pt]
  \includegraphics[width=0.9\linewidth]{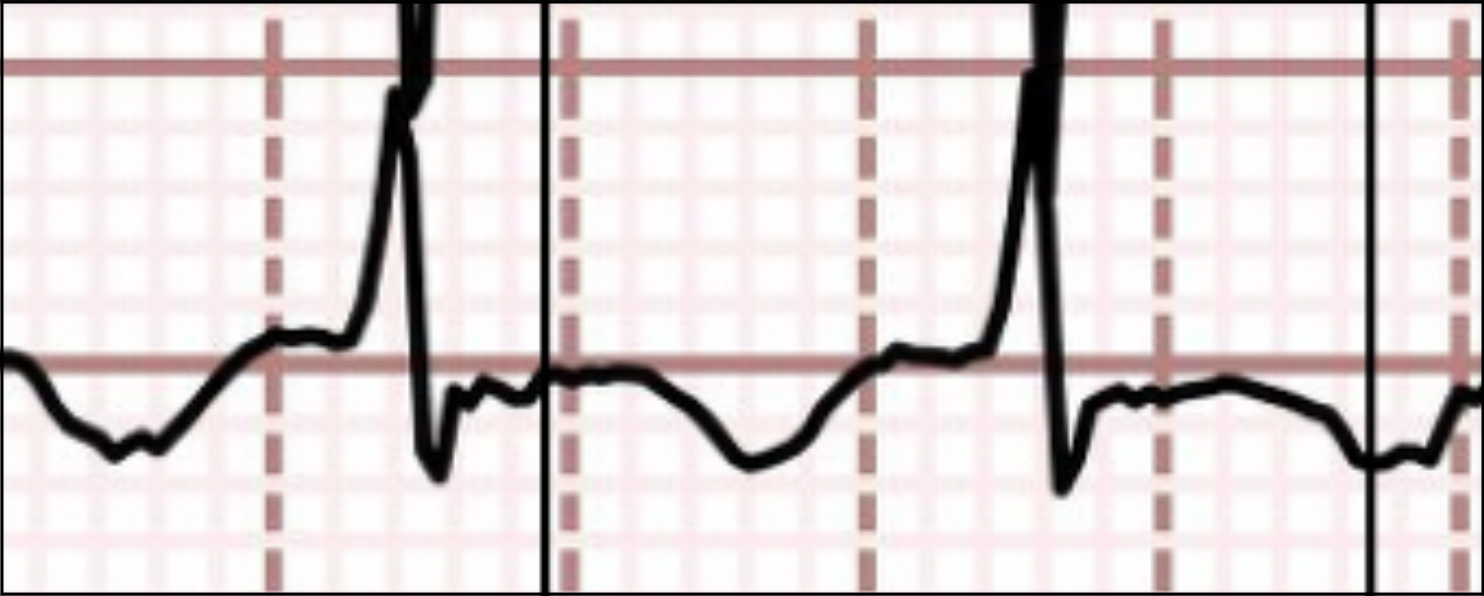} 
  \caption{Example of the augmentations. Upper image indicates OverlaySignal augmentation. Bottom image indicates the AddVerticalLines augmentation.}
  \label{fig:augmentation-examples}
\end{figure}


\subsubsection{Segmentation Evaluation}
The performance of the segmentation model was evaluated using two primary metrics: Intersection over Union (IoU) and the combined loss value described earlier.
The IoU, also known as the Jaccard index, is defined as:
$$ \text{IoU} = \frac{|Y \cap \hat{Y}|}{|Y \cup \hat{Y}|} = \frac{\text{TP}}{\text{TP} + \text{FP} + \text{FN}} $$
where TP, FP, and FN represent the number of true positive, false positive, and false negative pixels, respectively.

\subsection{Signal Digitization}
\label{sec:method-digitization}

The output of the segmentation stage is a binary mask (in PNG format) representing the ECG signal. This mask serves as the input for the digitization stage. The ground truth for this stage is the reference digital signal provided in the original dataset. In other words, this section aims to convert the middle image in Figure~\ref{fig:raw_dataset_sample} into  the bottom image in Figure~\ref{fig:raw_dataset_sample}. The number of samples for training, validation, and testing in this phase remains consistent with the segmentation stage. This section includes 

\subsubsection{Grid Detection}
The initial step in converting the segmented signal mask into clinically relevant voltage and time series is the detection of the underlying ECG grid. This process commences with the conversion of the input image to a grayscale representation. Given that ECG grid lines are typically rendered in a specific color, such as red, which translates to a distinct range of grayscale intensities, pixel intensity clustering is employed. The cluster associated with the median intensity values is generally chosen to isolate the grid lines, resulting in a preliminary binary grid mask. This mask is subsequently refined using a sequence of morphological operations, such as opening and closing, to eliminate noise and improve the coherence and definition of the grid lines. Following these enhancements, the Hough Line Transform, implemented in the OpenCV library \cite{opencv_library}, is applied to identify the dominant horizontal and vertical lines in the refined grid mask. The average pixel distance corresponding to one large grid square is then calculated for both horizontal ($width\_pixels$) and vertical ($height\_pixels$) dimensions by analyzing the spacing between these detected lines. In scenarios where grid lines are sparse, especially vertically, which can occur with small image crops that do not encompass multiple horizontal lines, a square grid is assumed as a heuristic. In such cases, $height\_pixels$ is set to be identical to $width\_pixels$.

\subsubsection{Signal Digitization}
Our methodology for converting the binary signal mask into a digital time series draws significant inspiration from the work of Fortune et al. \cite{article-23}, particularly their advanced techniques for signal detection and extraction, which are available in the \texttt{paper-ecg} software library \cite{digitize-github}. A critical first step in their approach is an adaptive thresholding process. This process begins by converting the input image to grayscale, after which an initial threshold is established using Otsu's method \cite{otsu79}. Otsu's method is designed to select a threshold that minimizes the intra-class variance of pixel intensities, effectively separating foreground from background.

Following the initial Otsu thresholding, a "hedging factor" is introduced, starting at a value of 1.0. The image is binarized using the product of the Otsu threshold and this hedging factor. This factor is then iteratively decreased by 5\% in each step, and the binarization is repeated, as long as grid lines remain detectable within the image. The iterative reduction of the hedging factor continues until either the grid lines are no longer detected or the factor reaches a predefined minimum of 0.6. This iterative refinement aims to selectively remove the lighter grid lines from the image while carefully preserving the darker pixels corresponding to the ECG signal itself. An optional simple denoising step can also be applied at the end of this stage to further clean the resulting binary image, which ideally contains only the ECG signal pixels.

Once the adaptive thresholding yields a clean binary representation of the ECG signal, the next stage is to extract the precise signal path. This is achieved by first invoking the described adaptive thresholding procedure and then employing Viterbi's algorithm \cite{Viterbi1967}. For each column in the binarized image, the algorithm identifies the centers of all contiguous regions of `True` (signal) pixels, treating these centers as potential nodes in a graph. A cost is associated with transitioning between nodes in adjacent columns; this cost is a weighted sum (with $\alpha = 0.5$) of the Euclidean distance between the nodes and the change in angle from the previous segment, promoting smoothness.

Dynamic programming is then utilized to determine the optimal, least-cost path from the leftmost column of the image to the rightmost. The sequence of y-coordinates derived from this optimal path constitutes the raw digitized signal. Should any columns in the image be devoid of signal pixels, interpolation is used to ensure a continuous signal. The underlying principle of this combined approach is that the progressive lowering of the threshold effectively isolates the ECG signal from the grid without resorting to complex filtering techniques. Subsequently, the application of Viterbi's algorithm guarantees the extraction of a continuous and smooth signal trace that is robust to localized breaks or noise in the mask.

\subsubsection{Post-processing}
With the grid dimensions ($width\_pixels$ and $height\_pixels$ per large grid square) established, and knowing the standard ECG scale (typically 0.5 mV per vertical large grid and 0.2 seconds per horizontal large grid), the raw pixel trace derived from the segmentation mask undergoes several post-processing steps. First, the y-coordinate of the signal (often taken as the centerline of the binary mask) is extracted for each x-coordinate. This raw pixel-based signal is then resampled to a standard sampling rate, such as 100 Hz, using interpolation. This resampling is guided by the detected horizontal grid scale, which represents the time axis, ensuring that the digitized signal has a consistent temporal resolution. Subsequently, to achieve optimal alignment between the extracted signal and the ground truth, a cross-correlation analysis is conducted. This step effectively identifies and corrects for minor horizontal shifts, or lags, which are commonly observed to be in the range of 3 to 5 samples. Finally, to address potential baseline wander or DC offset discrepancies between the predicted signal and the reference, the median value of the predicted signal is subtracted from all its samples. This baseline removal normalizes the signal for accurate amplitude-based comparisons, although it is important to note that this step does not influence the Pearson correlation coefficient. Collectively, these post-processing adjustments correct for horizontal (temporal) and vertical (amplitude) biases, thereby enabling a more precise and reliable comparison against the ground truth signal.

\subsubsection{Evaluation Metrics}
The quality of the final digitized ECG signal is assessed using two standard metrics: Mean Squared Error (MSE) and Pearson Correlation Coefficient ($\rho$).
\begin{itemize}
    \item \textbf{Mean Squared Error (MSE)}: Measures the average squared difference between the predicted signal ($S_{pred}$) and the ground truth signal ($S_{gt}$), each with $N$ samples.
    $$ \text{MSE} = \frac{1}{N} \sum_{i=1}^{N} (S_{gt,i} - S_{pred,i})^2 $$
    A lower MSE indicates a better match in terms of signal amplitude and shape.

    \item \textbf{Pearson Correlation Coefficient ($\rho$)}: Measures the linear correlation between the predicted and ground truth signals.
    $$ \rho(S_{gt}, S_{pred}) = \frac{\text{cov}(S_{gt}, S_{pred})}{\sigma_{S_{gt}} \sigma_{S_{pred}}} $$
    $$ = \frac{\sum_{i=1}^{N} (S_{gt,i} - \bar{S}_{gt})(S_{pred,i} - \bar{S}_{pred})}{\sqrt{\sum_{i=1}^{N} (S_{gt,i} - \bar{S}_{gt})^2} \sqrt{\sum_{i=1}^{N} (S_{pred,i} - \bar{S}_{pred})^2}} $$
    where $\bar{S}$ denotes the mean of the signal. A value closer to 1 indicates a strong positive linear relationship.
\end{itemize}

%% file: results.tex
This section details the experimental setup, presents the results obtained for both the ECG signal segmentation and digitization stages, and provides a discussion of these outcomes. We first evaluate two distinct approaches for ECG signal segmentation, followed by an analysis of the digitization process. The performance of our proposed digitization pipeline is then compared against a previously published method by Tereshchenkolab et al. \cite{article-23}.

\subsection{ECG Signal Segmentation}
Accurate segmentation of the ECG signal from the paper background is crucial for reliable digitization. We explored two different neural network architectures for this task: a YOLO-based model and a U-Net-based model.

\subsubsection{Segmentation using YOLO-based Architecture}
Our initial approach to ECG signal segmentation involved utilizing a model based on the You Only Look Once (YOLO) architecture, accessed via the Ultralytics library \cite{ultralytics-ref, yolov7}. YOLO models are primarily known for object detection but have been extended to instance segmentation tasks, typically by predicting masks for each detected bounding box.

The model was trained on the \texttt{segmentation-dataset} derived from the introduced dataset \cite{dataset-paper,dataset-zenodo}. A notable characteristic of this dataset is the intentional reduction of vertical spacing between ECG leads in some samples, thereby increasing the occurrence of signal overlaps and providing a richer dataset for training models robust to such conditions. Training was conducted in the Google Colab environment using a T4 GPU.

The YOLO-based model, based on \texttt{yolo11n}, was trained for up to 200 epochs, leveraging pretrained weights and an 'auto' optimizer setting. Input images were resized to 640×640 pixels and processed with a batch size of 16. The learning rate schedule featured an initial (\texttt{lr0}) learning rate of 0.01, a momentum of 0.937, and a weight decay of 0.0005. This was preceded by a three-epoch warmup phase, using a warmup momentum of 0.8 and a warmup bias learning rate of 0.1. The composite loss function incorporated specific gains for different components: 7.5 for box regression (\texttt{box}), 0.5 for classification (\texttt{cls}) – although the primary task was binary signal segmentation and 1 for dice loss. Notably, Non-Maximum Suppression (NMS) was disabled for this segmentation setup. Standard data augmentations were applied during training, including HSV color space adjustments (hue variation of ±0.015, saturation variation of ±0.7, and value variation of ±0.4), random translations (by a factor of ±0.1), random scaling (by a factor of ±0.5), along with vertical flips (with a probability of 0.5) and horizontal flips (with a probability of 0.3). Early stopping was active, which concluded the training process after 117 epochs.

\begin{figure}[t]
  \centering
  \includegraphics[width=0.9\linewidth]{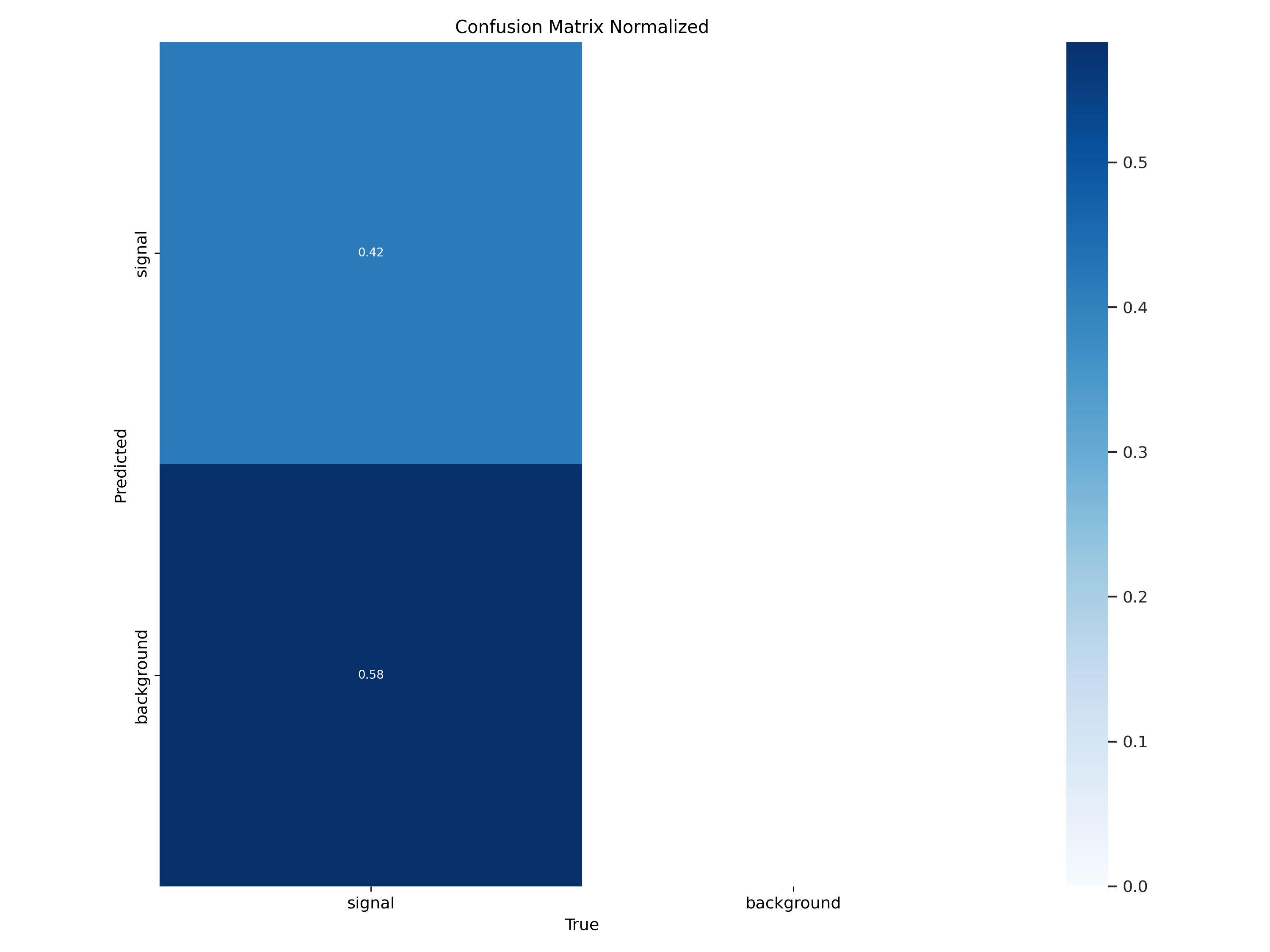} 
  \caption{Confusion matrix for the YOLO-based segmentation model on the test set.}
  \label{fig:yolo_confusion_matrix}
\end{figure}

Despite these extensive configurations and multiple training attempts, the YOLO-based model did not yield satisfactory results. As illustrated in the confusion matrix (Figure~\ref{fig:yolo_confusion_matrix}), a significant number of samples (58\% of instances in this particular dataset) had no signal region detected by the model.

\begin{figure}[t]
  \centering
  \includegraphics[width=0.9\linewidth]{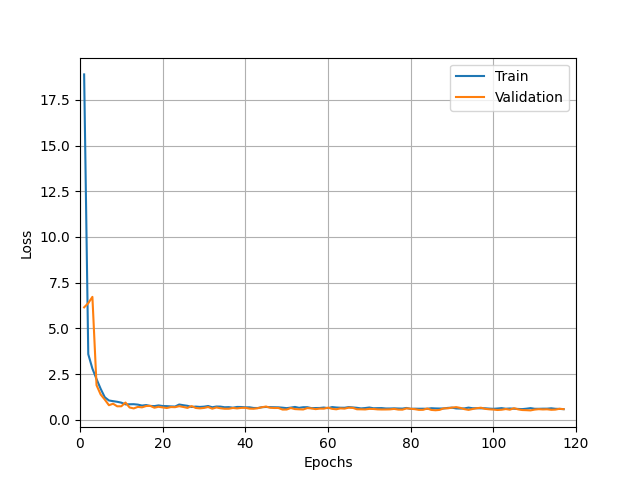} 
  \includegraphics[width=0.9\linewidth]{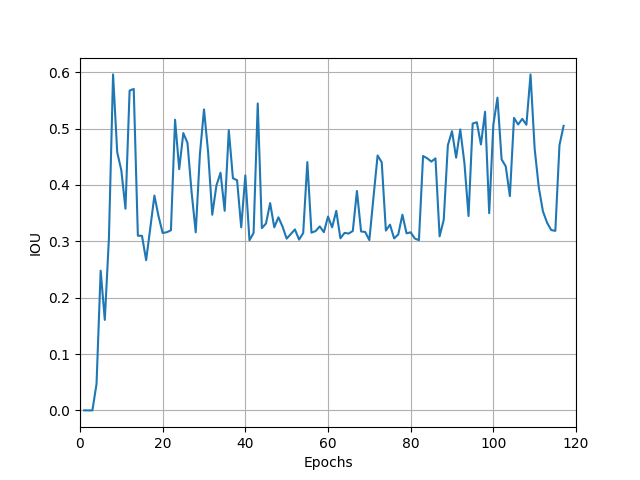} 
  \caption{Training and validation loss (upper) and validation IoU (bottom) for the YOLO-based segmentation model.}
  \label{fig:yolo_loss_iou} 
  \label{fig:yolo_loss} 
  \label{fig:yolo_iou} 
\end{figure}

The training and validation loss curves are presented in Figure~\ref{fig:yolo_loss}, and the IoU on the validation set is shown in Figure~\ref{fig:yolo_iou}. The IoU remained low, reaching a maximum of approximately $0.6$. This performance was achieved even without applying the more complex custom augmentations described in Section \ref{sec:method-signal-segmentation}, where one might expect the model to overfit to the training data more easily. The poor performance, particularly the failure to detect signals in many instances, indicated that this architecture was not well-suited for the fine-grained segmentation required for ECG traces.

\subsubsection{Segmentation using U-Net Architecture}
Given the challenges with the YOLO-based approach, we employed a U-Net architecture \cite{Ronneberger2015} with a ResNet34 \cite{He2016} backbone, as detailed in Section \ref{sec:method-signal-segmentation}. The model was trained on the same \texttt{segmentation-dataset} \cite{dataset-paper, dataset-zenodo} and in the same Google Colab T4 GPU environment.

\begin{figure}[t]
  \centering
  \includegraphics[width=0.9\linewidth]{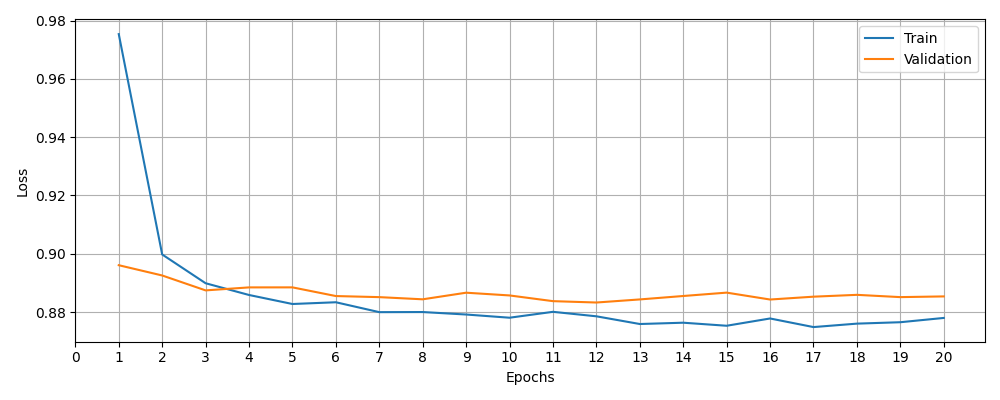} 
  \includegraphics[width=0.9\linewidth]{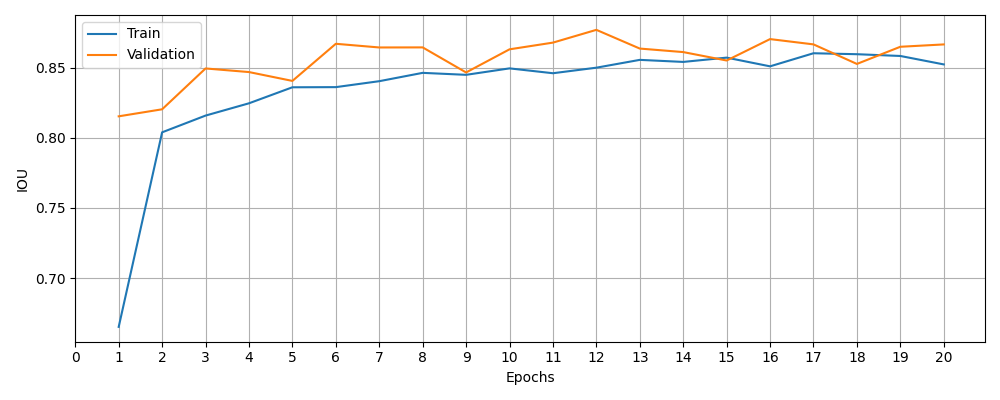} 
  \caption{Training and validation loss (up) and Intersection over Union (IoU) (bottom) for the U-Net based segmentation model.}
  \label{fig:unet_loss_iou}
\end{figure}

The training progress for the U-Net model is illustrated in Figure~\ref{fig:unet_loss_iou}. The loss curves for both training and validation sets show a consistent decrease, while the IoU scores for both sets demonstrate a steady improvement, reaching a maximum of $0.87$ on the validation set. Importantly, the validation loss and IoU curves closely follow the training curves without significant divergence, indicating no overfitting during the 20 epochs of training.

\begin{figure*}[t]
  \centering
  \includegraphics[width=0.9\linewidth]{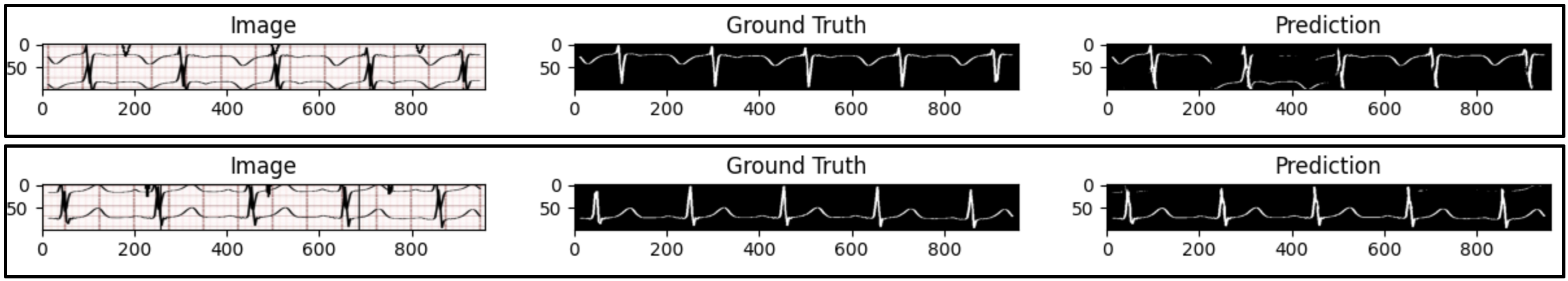} 
  \caption{Qualitative results of the U-Net segmentation model on a sample with overlapping signals. Left column: Input image. Middle column: Ground truth mask. Right column: Predicted mask. Top row: Epoch 3. Bottom row: Epoch 19.}
  \label{fig:unet_segmentation_progress}
\end{figure*}

Figure~\ref{fig:unet_segmentation_progress} showcases the qualitative improvement of the U-Net model's predictions on a challenging sample containing overlapping signals. At an early stage of training (epoch 3), the model incorrectly includes a portion of an interfering signal from the bottom of the image. However, by the final epoch (epoch 19), the model has learned to correctly delineate the primary ECG trace, effectively ignoring the overlapping segment. This demonstrates the model's ability to learn complex signal boundaries, benefiting from the architecture and the custom augmentations designed for such scenarios.

\subsection{Signal Digitization}
Following successful segmentation, the resulting binary mask of the ECG signal is processed to extract a time-series representation. This subsection details the performance of our digitization pipeline and compares it with the baseline results.

\subsubsection{Baseline Method(Tereshchenkolab et al. \cite{article-23})}
We implemented and tested the digitization algorithm described in \cite{article-23}, available from their public repository \cite{digitize-github}. This method was applied to single-lead ECG images from our test set. To ensure a fair comparison with our proposed method, we applied similar post-processing steps to the output of \cite{article-23}. pipeline. Specifically, the raw digitized signal was resampled to a standard 100 Hz sampling rate using interpolation, guided by an estimated horizontal grid scale. A cross-correlation analysis was then performed to correct for minor horizontal shifts (lags), typically observed to be between 3 to 5 samples. Finally, baseline wander was addressed by subtracting the median value of the predicted signal from all its samples.

\subsubsection{Proposed Digitization Method}
\begin{figure}[t]
  \centering
  \includegraphics[width=0.9\linewidth]{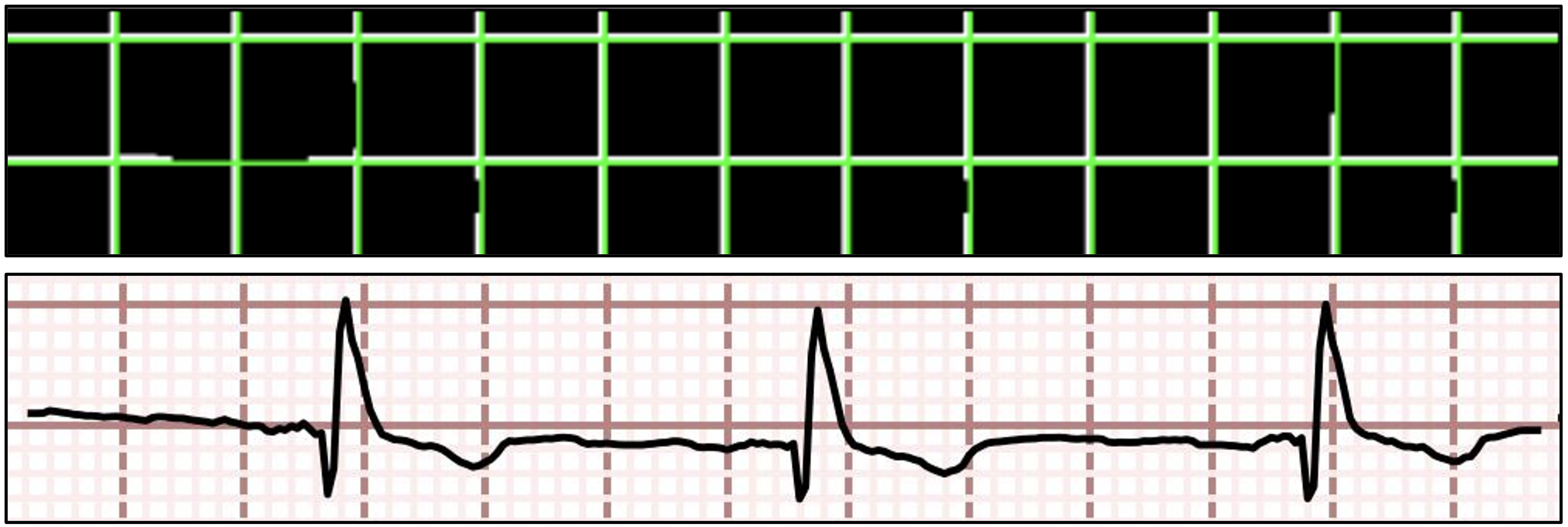} 
  \caption{Example of detected grid lines (green lines in upper image) overlaid on an input ECG lead image (bottom).}
  \label{fig:grid_detection_example}
\end{figure}

Our digitization process, as described in Section \ref{sec:method-digitization}, begins with grid detection. The input color image is used for this step to maximize the visibility of grid lines. An example of a detected grid overlaid on an ECG image is shown in Figure~\ref{fig:grid_detection_example}. The detected horizontal and vertical grid lines (shown in green) allow for the calculation of pixel distances per standard ECG grid square. In cases where insufficient grid lines were present in a single lead image for robust estimation (e.g., too few horizontal lines), we heuristically assumed a square grid, setting $height\_pixels = width\_pixels$.

\begin{figure}[t]
  \centering
  \includegraphics[width=0.9\linewidth]{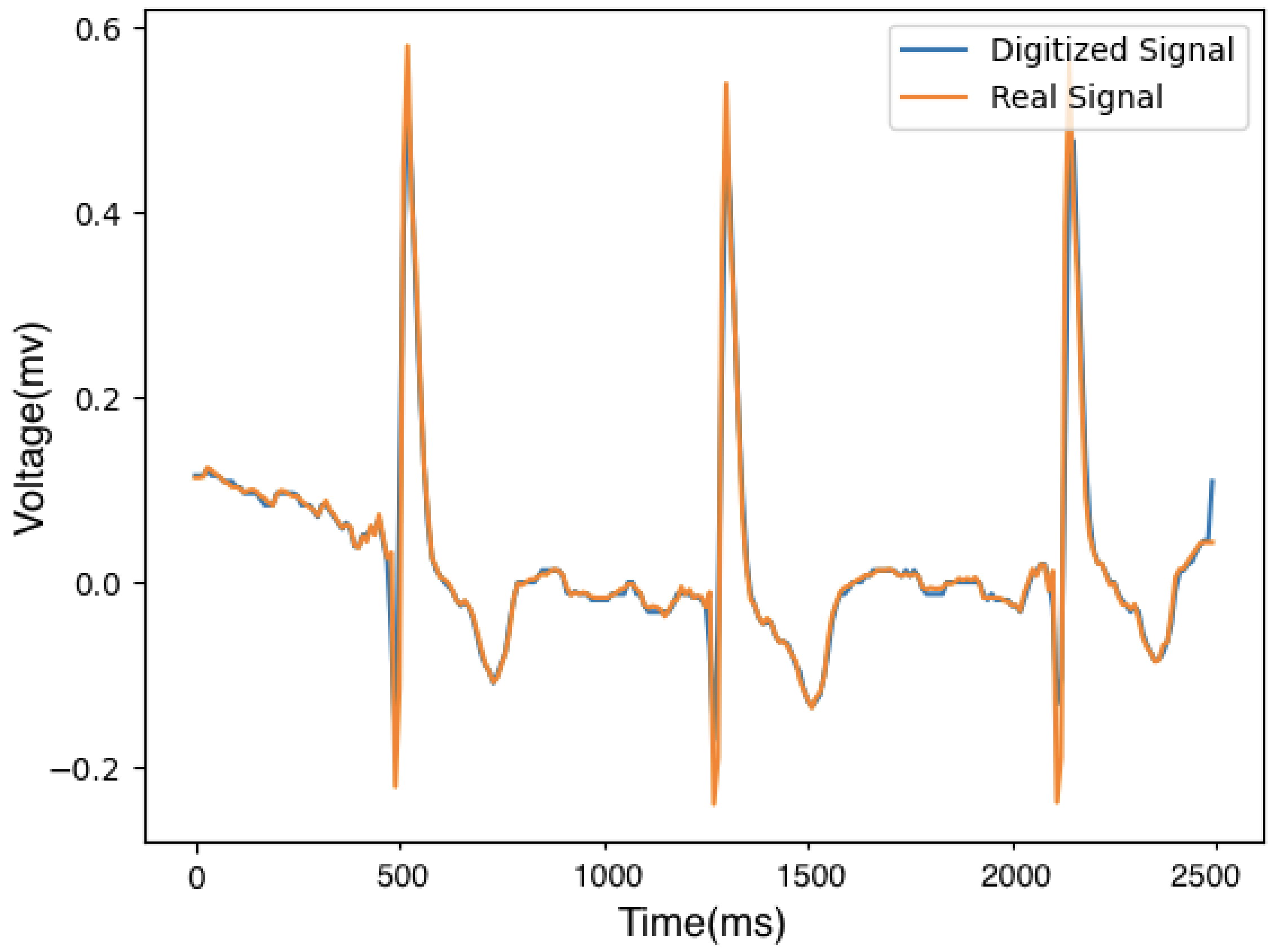} 
  \caption{Example of a digitized ECG signal from our proposed method (e.g., blue line) compared against the ground truth signal (e.g., red line).}
  \label{fig:digitized_signal_example}
\end{figure}

The binary mask produced by our U-Net segmentation model serves as the input for the signal extraction component of the digitization stage. The combination of adaptive thresholding and Viterbi path finding, followed by the post-processing steps (resampling, lag correction, and baseline removal as detailed in Section 2.3.3), yields the final time-series data. Figure~\ref{fig:digitized_signal_example} illustrates an example of a digitized signal generated by our pipeline, plotted alongside the ground truth signal, demonstrating a high degree of concordance.

\subsubsection{Comparative Evaluation}
\label{sec:comparative-evaluation}
We evaluated our proposed digitization method against the Tereshchenkolab et al. \cite{article-23} method using a subset of the dataset available at \cite{dataset-zenodo}. This subset, named \texttt{overlap} data, was divided into two groups: 100 samples manually verified to contain signal overlaps and 185 samples manually verified to have no signal overlaps. The performance was assessed using Mean Squared Error (MSE) and Pearson Correlation Coefficient ($\rho$).

As indicated by the calculated metrics in Table~\ref{tab:results_no_overlap} and Table~\ref{tab:results_overlap}, our proposed method achieved superior performance in terms of both lower MSE and higher Pearson correlation across both data categories (with and without signal overlap) when compared to the Tereshchenkolab et al. \cite{article-23} method applied to our dataset. This suggests that the combination of our U-Net based segmentation, tailored grid detection, and signal extraction pipeline is more robust, particularly in handling challenging cases with overlapping signals.

\begin{table}[t]
  \centering
  \caption{Digitization performance on samples \textbf{without} signal overlap (N=185).}
  \label{tab:results_no_overlap}
  \begin{tabular}{lcc}
    \toprule
    Method & Tereshchenkolab et al. \cite{article-23} & Proposed Method \\
    \midrule
    MSE ($\downarrow$) & 0.0015 & 0.0010 \\
    MSE Std ($\downarrow$) & 0.0014 & 0.0008 \\
    MSE Max ($\downarrow$) & 0.0086 & 0.0059 \\
    $\rho$ ($\uparrow$) & 0.9366 & 0.9644 \\
    $\rho$ Min ($\uparrow$) & 0.7838 & 0.8572 \\
    $\rho$ Std ($\uparrow$) & 0.0519 & 0.0205 \\
    \bottomrule
  \end{tabular}
\end{table}

\begin{table}[t]
  \centering
  \caption{Digitization performance on samples \textbf{with} signal overlap (N=100).}
  \label{tab:results_overlap}
  \begin{tabular}{lcc}
    \toprule
    Method & Tereshchenkolab et al. \cite{article-23} & Proposed Method \\
    \midrule
    MSE ($\downarrow$) & 0.0178 & 0.0029 \\
    MSE Std ($\downarrow$) & 0.0556 & 0.0032 \\
    MSE Max ($\downarrow$) & 0.4375 & 0.0216 \\
    $\rho$ ($\uparrow$) & 0.8676 & 0.9641 \\
    $\rho$ Min ($\uparrow$) & 0.1897 & 0.6256 \\
    $\rho$ Std ($\uparrow$) & 0.1477 & 0.0812 \\
    \bottomrule
  \end{tabular}
\end{table}

%% file: discussion.tex
In this study, we proposed a deep learning-based digitization pipeline for single-lead ECG images, designed specifically to address the challenge of overlapping signals. Using a U-Net architecture trained on a custom dataset, which was synthetically enriched with numerous examples of overlapping signals and custom data augmentations, we achieved a high Pearson correlation of approximately $0.96$ for both non-overlapping and challenging overlapping cases. This result marks a significant improvement upon a recent prominent study where digitization correlation was reported to fall into the 60--70\% range when signal overlaps were present \cite{article-21}. Moreover, our method consistently outperformed a well-established baseline technique across MSE and correlation metrics. The performance gap was particularly substantial for the challenging overlapping signals. 

The core of our approach lies in accurately isolating the primary ECG trace from its background and any interfering elements, particularly other ECG signals. This was achieved by training a U-Net based segmentation network on a dataset specifically curated to include a higher proportion of samples with signal overlap. Furthermore, we introduced custom data augmentation techniques, such as \texttt{OverlaySignal}, to enhance the model's resilience to such scenarios. This targeted training strategy proved crucial for the segmentation model's success. Subsequently, the generated binary mask was converted into a time-series signal using established signal processing techniques, importantly preceded by an automated grid detection module. This grid detection step ensures the adaptability of our digitization process to various ECG paper scales and scanning resolutions, a critical factor for practical applicability. The primary research question we aimed to answer was how to achieve robust digitization in the presence of signal overlaps. Our findings suggest that this can be effectively addressed by training a specialized segmentation network on data representative of this challenge, significantly amplified by tailored augmentations.

Our choice of the U-Net architecture for the segmentation task was validated by its significantly superior performance compared to a YOLO-based approach. Several factors contribute to this difference. Firstly, the nature of the task, which involves precise pixel-level delineation of thin ECG traces, aligns well with U-Net's design, originally conceived for biomedical image segmentation \cite{Ronneberger2015}. YOLO, while capable of instance segmentation, is fundamentally optimized for bounding box detection and may struggle with the highly irregular and fine-grained shapes of ECG signals. Secondly, U-Net's characteristic skip connections, which propagate feature information from the encoder to the decoder, are crucial for preserving the fine details of the thin ECG lines through the multiple down-sampling and up-sampling stages. In contrast, aggressive down-sampling in YOLO architectures without such meticulous feature restoration can lead to the loss of these narrow signal components. Lastly, U-Net benefits from dense supervision, where every pixel in the output mask contributes to the loss function, providing a rich training signal. YOLO's supervision for segmentation masks, often tied to initial bounding box proposals, might offer a sparser learning signal for this specific type of task.

The efficacy of our complete pipeline was benchmarked against the method proposed by Tereshchenkolab et al. \cite{article-23}, a notable prior work in this domain. As detailed in Section~\ref{sec:comparative-evaluation}'s results (Tables~\ref{tab:results_no_overlap} and \ref{tab:results_overlap}), our proposed method demonstrated superior performance across both non-overlapping and overlapping ECG samples from the \texttt{overlap} dataset derived from \cite{dataset-paper,dataset-zenodo}. For non-overlapping signals, our method achieved a lower Mean Squared Error (MSE) of $0.0010$ compared to $0.0015$ and a higher Pearson Correlation Coefficient ($\rho$) of $0.9644$ versus $0.9366$. The improvement was even more pronounced for challenging samples with signal overlap. Here, our method yielded an MSE of $0.0029$ in contrast to $0.0178$ for the baseline, and a $\rho$ of $0.9641$ compared to $0.8676$. These quantitative results strongly indicate the enhanced robustness of our approach, particularly its ability to accurately digitize signals even when they are partially obscured by others.
Furthermore, when comparing our results to other deep learning-based digitization efforts, such as the work by Wu et al. \cite{article-21}, who reported average correlations of 60-70\% for some leads in 12-by-1 ECGs and 80-90\% for 3-by-1 ECGs without specific exclusions for overlap, our achieved correlation exceeding $0.96$ on a the images with overlapping signals. This underscores the effectiveness of our dedicated strategy for mitigating the impact of signal interference.

The strengths of our proposed method are multifaceted. Foremost is its enhanced capability in handling ECGs with overlapping signals, a direct result of the specialized segmentation model and training strategy. Secondly, the integrated grid detection module imparts crucial adaptability, allowing the system to process ECGs scanned at various resolutions or printed on different grid scales without manual calibration. Thirdly, the open-source nature of our implementation is intended to facilitate its adoption, validation, and extension by the broader research community, enabling rapid application to new datasets. Finally, the flexible data augmentation pipeline not only proved effective for the current challenges but also provides a framework for incorporating robustness to other conditions, such as the presence of handwriting or artifacts from different image acquisition processes.

Despite the promising results, particularly in handling overlapping signals, our study acknowledges certain limitations that pave the way for valuable future research. The current segmentation model, while demonstrating robustness, was developed using a dataset meticulously curated to feature a high incidence of signal overlaps—a strategy effective for targeting this specific, challenging artifact. To further enhance its real-world applicability and generalization, future efforts would greatly benefit from evaluating and potentially fine-tuning the model on a substantially larger and more diverse corpus of real-world paper ECGs. Such an ideal benchmark dataset would need to encompass samples from a myriad of clinical settings worldwide to capture inherent variabilities in paper types, printing formats (such as horizontal versus vertical layouts indicative of different institutional practices), and idiosyncratic elements like physician signatures or administrative notes. It should also include data acquired through a wide array of methods – ranging from professional scanners to common mobile phone cameras under diverse ambient lighting and perspective angles – to ensure robustness against varied input qualities, alongside a broader spectrum of naturally occurring artifacts such as stains, creases, tears, fading, and diverse styles of incidental handwriting, beyond the specifically curated overlaps. The generation of such a comprehensive dataset, while paramount for advancing the field, admittedly presents considerable logistical challenges related to sourcing physical records, the intensive manual effort and expertise required for precise ground truth digitization and annotation, and the financial investment needed for high-quality scanning or printing and subsequent data processing.

Beyond dataset scope, while current augmentations address some forms of noise, dedicated strategies, potentially through novel augmentation techniques, are warranted to improve robustness against extensive handwriting directly superimposed on or immediately adjacent to the ECG trace. Moreover, the current pipeline's focus on pre-cropped single leads would necessitate the research and integration of an accurate and efficient automated lead detection and separation module to evolve into a fully automated system capable of processing complete multi-lead ECG sheets. Similarly, our rule-based grid detection, while adaptive, could be advanced by exploring data-driven approaches, perhaps by incorporating grid line parameter estimation as an auxiliary task within the U-Net architecture itself, potentially leading to a more robust and seamlessly integrated solution for geometric calibration. Finally, a critical future direction remains the comprehensive clinical validation of the digitized ECG signals, involving rigorous assessment by medical professionals to ascertain their diagnostic quality and reliability for meaningful application in real-world clinical decision-making or research endeavors.

%% file: conclusion.tex
This paper addressed the persistent challenge of accurately digitizing ECG image recordings, with a specific focus on robustly handling single leads compromised by signal overlaps—a prevalent and often overlooked issue in existing methodologies. Our proposed two-stage pipeline directly tackles this limitation. The first stage employs a U-Net-based segmentation network, rigorously trained on an enriched dataset with overlapping signals and fortified by custom data augmentations, to precisely isolate the primary ECG trace. The subsequent stage converts this refined binary mask into a time-series signal using established digitization techniques, further enhanced by an adaptive grid detection module to ensure versatility across diverse ECG formats and scales.

Our primary contribution is the demonstration of an effective strategy to significantly improve digitization accuracy, especially in the presence of signal overlaps. By specifically tailoring the training data and augmentation procedures for the segmentation network, we successfully mitigated the adverse effects of interfering signals, leading to a more reliable extraction of the true ECG waveform. This approach yielded superior performance compared to existing methods, such as that by Tereshchenkolab et al. \cite{article-23}, particularly on ECG samples exhibiting such overlaps. Ultimately, the presented methodology offers a significant advancement in the automated digitization of challenging paper ECGs, establishing a robust foundation for converting analog ECG records into analyzable digital data, thus enhancing the utility of historical medical information for contemporary research and clinical applications.

%% file: article.bbl
\begin{thebibliography}{10}

\bibitem{birnbaum2014role}
Y.~Birnbaum and L.~Jackson, ``The role of the 12-lead electrocardiogram in the
  current era of cardiovascular imaging,'' {\em Journal of Electrocardiology},
  vol.~47, no.~3, pp.~215--225, 2014.

\bibitem{adedinsewo2022digitizing}
D.~A. Adedinsewo, H.~Siddiqui, P.~W. Johnson, E.~J. Douglass, M.~Cohen-Shelly,
  Z.~I. Attia, P.~Friedman, P.~A. Noseworthy, and R.~E. Carter, ``Digitizing
  paper based ecg files to foster deep learning based analysis of existing
  clinical datasets: An exploratory analysis,'' {\em Intelligence-Based
  Medicine}, vol.~6, p.~100070, 2022.

\bibitem{reyna2024ecg}
M.~A. Reyna, J.~Weigle, Z.~Koscova, K.~Campbell, K.~K. Shivashankara,
  S.~Saghafi, S.~Nikookar, M.~Motie-Shirazi, Y.~Kiarashi, S.~Seyedi, {\em
  et~al.}, ``Ecg-image-database: A dataset of ecg images with real-world
  imaging and scanning artifacts; a foundation for computerized ecg image
  digitization and analysis,'' {\em arXiv preprint arXiv:2409.16612}, 2024.

\bibitem{lence2023automatic}
A.~Lence, F.~Extramiana, A.~Fall, J.-E. Salem, J.-D. Zucker, and E.~Prifti,
  ``Automatic digitization of paper electrocardiograms--a systematic review,''
  {\em Journal of Electrocardiology}, vol.~80, pp.~125--132, 2023.

\bibitem{article-12}
S.~Mallawaarachchi, M.~P.~N. Perera, and N.~D. Nanayakkara, ``Toolkit for
  extracting electrocardiogram signals from scanned trace reports,'' in {\em
  2014 IEEE Conference on Biomedical Engineering and Sciences (IECBES)},
  pp.~868--873, 2014.

\bibitem{article-1}
Y.~Li, Q.~Qu, M.~Wang, L.~Yu, J.~Wang, L.~Shen, and K.~He, ``Deep learning for
  digitizing highly noisy paper-based ecg records,'' {\em Computers in Biology
  and Medicine}, vol.~127, p.~104077, 2020.

\bibitem{article-22}
A.~Demolder, V.~Kresnakova, M.~Hojcka, V.~Boza, A.~Iring, A.~Rafajdus,
  S.~Rovder, T.~Palus, M.~Herman, F.~Bauer, V.~Jurasek, R.~Hatala, J.~Bartunek,
  B.~Vavrik, and R.~Herman, ``High precision ecg digitization using artificial
  intelligence,'' {\em Journal of Electrocardiology}, vol.~90, p.~153900, 2025.

\bibitem{article-20}
J.~Wang, Y.~Pang, Y.~He, and J.~Pan, ``Ecg waveform extraction from paper
  records,'' in {\em Image and Graphics} (Y.~Zhao, X.~Kong, and D.~Taubman,
  eds.), (Cham), pp.~505--512, Springer International Publishing, 2017.

\bibitem{article-14}
R.~Patil and R.~Karandikar, ``Image digitization of discontinuous and degraded
  electrocardiogram paper records using an entropy-based bit plane slicing
  algorithm,'' {\em Journal of Electrocardiology}, vol.~51, no.~4,
  pp.~707--713, 2018.

\bibitem{article-11}
S.~Lobodzinski, U.~Teppner, and M.~Laks, ``State of the art techniques for
  preservation and reuse of hard copy electrocardiograms,'' {\em Journal of
  electrocardiology}, vol.~36 Suppl, pp.~151--5, 02 2003.

\bibitem{article-23}
J.~D. Fortune, N.~E. Coppa, K.~T. Haq, H.~Patel, and L.~G. Tereshchenko,
  ``Digitizing ecg image: A new method and open-source software code,'' {\em
  Computer Methods and Programs in Biomedicine}, vol.~221, p.~106890, 2022.

\bibitem{article-19}
G.~Waits and E.~Z. Soliman, ``Digitizing paper electrocardiograms: Status and
  challenges,'' {\em Journal of Electrocardiology}, vol.~50, 09 2016.

\bibitem{article-5}
X.~Yu, Y.~Huang, J.~Wu, J.~Wang, and W.~Cai, ``From paper to digital: Ecg
  processing with u-net digitization and resnet classification,'' 12 2024.

\bibitem{article-7}
R.~Silva, Y.~Yang, M.~Morier, S.~Al-Ali, and M.~Sermesant, ``Your-lead: Yolo
  and u-net for reconstruction of ecg lead signals,'' 12 2024.

\bibitem{article-15}
R.~Patil, B.~Narkhede, S.~Varma, S.~Suraliya, and N.~Mehendale, ``Auto lead
  extraction and digitization of ecg paper records using cgan,'' 2022.

\bibitem{article-21}
H.~Wu, K.~H.~K. Patel, X.~Li, B.~Zhang, C.~Galazis, N.~Bajaj, A.~Sau, X.~Shi,
  L.~Sun, Y.~Tao, {\em et~al.}, ``A fully-automated paper ecg digitisation
  algorithm using deep learning,'' {\em Scientific Reports}, vol.~12, no.~1,
  p.~20963, 2022.

\bibitem{article-13}
S.~Mishra, G.~Khatwani, R.~Patil, D.~Sapariya, V.~Shah, D.~Parmar, S.~Dinesh,
  P.~Daphal, and N.~Mehendale, ``Ecg paper record digitization and diagnosis
  using deep learning,'' {\em Journal of Medical and Biological Engineering},
  vol.~41, 06 2021.

\bibitem{dataset-paper}
M.~Rahimi, R.~Karbasi, and A.-H. Vahabie, ``An open-source python framework and
  synthetic ecg image datasets for digitization, lead and lead name detection,
  and overlapping signal segmentation,'' 2025.

\bibitem{article-16}
R.~Patil and R.~Karandikar, ``Robust algorithm for digitization of degraded
  electrocardiogram paper records,'' {\em ICTACT Journal on Communication
  Technology}, vol.~8, pp.~1604--1609, 09 2017.

\bibitem{article-8}
A.~Isabel, G.~Jimenez-Perez, O.~Camara, and E.~Silva, ``Mobile app for the
  digitization and deep-learning-based classification of electrocardiogram
  printed records,'' in {\em 2021 Computing in Cardiology (CinC)}, vol.~48,
  pp.~1--4, 2021.

\bibitem{article-3}
F.~Badilini, T.~Erdem, W.~Zareba, and A.~J. Moss, ``Ecgscan: a method for
  conversion of paper electrocardiographic printouts to digital
  electrocardiographic files,'' {\em Journal of Electrocardiology}, vol.~38,
  no.~4, pp.~310--318, 2005.

\bibitem{article-6}
M.~Verlyck, J.~Dillon, S.~Creamer, and D.~Zhao, ``Wavie: A modular and
  open-source python implementation for fully automated digitisation of paper
  electrocardiograms,'' 12 2024.

\bibitem{article-9}
D.~Garg, D.~Thakur, S.~Sharma, and S.~Bhardwaj, ``Ecg paper records
  digitization through image processing techniques,'' {\em International
  Journal of Computer Applications}, vol.~48, 06 2012.

\bibitem{article-18}
T.~Tabassum and M.~Ahmad, ``Numerical data extraction from ecg paper recording
  using image processing technique,'' in {\em 2020 11th International
  Conference on Electrical and Computer Engineering (ICECE)}, pp.~355--358,
  2020.

\bibitem{Ronneberger2015}
O.~Ronneberger, P.~Fischer, and T.~Brox, ``U-net: Convolutional networks for
  biomedical image segmentation,'' {\em CoRR}, vol.~abs/1505.04597, 2015.

\bibitem{He2016}
K.~He, X.~Zhang, S.~Ren, and J.~Sun, ``Deep residual learning for image
  recognition,'' {\em CoRR}, vol.~abs/1512.03385, 2015.

\bibitem{Iakubovskii-2019}
P.~Iakubovskii, ``Segmentation models pytorch.''
  \url{https://github.com/qubvel/segmentation_models.pytorch}, 2019.

\bibitem{Falcon_PyTorch_Lightning_2019}
W.~Falcon and {The PyTorch Lightning team}, ``{PyTorch Lightning},'' Mar. 2019.

\bibitem{adam-optimizer}
D.~Kingma and J.~Ba, ``Adam: A method for stochastic optimization,'' {\em
  International Conference on Learning Representations}, 12 2014.

\bibitem{opencv_library}
G.~Bradski, ``{The OpenCV Library},'' {\em Dr. Dobb's Journal of Software
  Tools}, 2000.

\bibitem{digitize-github}
J.~D. Fortune, N.~E. Coppa, and L.~G. Tereshchenko, ``{paper-ecg},'' June 2021.

\bibitem{otsu79}
N.~Otsu, ``A threshold selection method from gray-level histograms,'' {\em IEEE
  Transactions on Systems, Man, and Cybernetics}, vol.~9, no.~1, pp.~62--66,
  1979.

\bibitem{Viterbi1967}
A.~Viterbi, ``Error bounds for convolutional codes and an asymptotically
  optimum decoding algorithm,'' {\em IEEE Transactions on Information Theory},
  vol.~13, no.~2, pp.~260--269, 1967.

\bibitem{ultralytics-ref}
G.~Jocher, J.~Qiu, and A.~Chaurasia, ``{Ultralytics YOLO},'' Jan. 2023.

\bibitem{yolov7}
C.-Y. Wang, A.~Bochkovskiy, and H.-Y.~M. Liao, ``Yolov7: Trainable
  bag-of-freebies sets new state-of-the-art for real-time object detectors,''
  2022.

\bibitem{dataset-zenodo}
M.~Rahimi, R.~Karbasi, and A.~hossein Vahabie, ``Multiple synthetic ecg image
  datasets for digitization, lead region and lead name detection, and signal
  segmentation,'' May 2025.

\end{thebibliography}
